\documentclass[runningheads]{llncs}
\usepackage{graphicx}
\usepackage{comment}
\usepackage{amsmath,amssymb} 
\usepackage{color}
\usepackage{subfig}
\usepackage{diagbox}
\renewcommand\tablename{Table S}
\renewcommand\thetable{\unskip\arabic{table}}

\usepackage[width=122mm,left=12mm,paperwidth=146mm,height=193mm,top=12mm,paperheight=217mm]{geometry}

\newcommand\blfootnote[1]{%
  \begingroup
  \renewcommand\thefootnote{}\footnote{#1}%
  \addtocounter{footnote}{-1}%
  \endgroup
}

\begin{document}
\pagestyle{headings}
\mainmatter

\title{Improving out-of-distribution generalization via multi-task self-supervised pretraining} 

\titlerunning{Improving O.O.D. generalization via multi-task self-supervised pretraining}
%
\author{Isabela Albuquerque\inst{1,*} \and
Nikhil Naik\inst{2} \and
Junnan Li\inst{2} \and Nitish Keskar\inst{2} \and Richard Socher\inst{2}}
\authorrunning{Albuquerque et al.}
%
\institute{INRS-EMT, Universit\'e du Qu\'ebec 
\and
Salesforce Research \\
}

\maketitle

\begin{abstract}
Self-supervised feature representations have been shown to be useful for supervised classification, few-shot learning, and adversarial robustness. We show that features obtained using self-supervised learning are comparable to, or better than, supervised learning for domain generalization in computer vision. We introduce a new self-supervised pretext task of predicting responses to Gabor filter banks and demonstrate that multi-task learning of compatible pretext tasks improves domain generalization performance as compared to training individual tasks alone. Features learnt through self-supervision obtain better generalization to unseen domains when compared to their supervised counterpart when there is a larger domain shift between training and test distributions and even show better localization ability for objects of interest. Self-supervised feature representations can also be combined with other domain generalization methods to further boost performance. 

\keywords{Self-supervised learning, out-of-distribution generalization, transfer learning.}
\end{abstract}

\section{Introduction}
Deep learning methods obtain impressive results on supervised learning benchmarks in computer vision, but struggle when tested on data distributions unseen during training time. This is not surprising since these models are optimized with empirical risk minimization (ERM) with the assumption that the examples from training and test sets are independently and identically drawn from the same distribution. However, machine learning models are often required to deal with a shift in data distribution or even with unseen distributions. Generalization to unseen distributions is important for building robust machine learning models. This problem is formally defined as the \textit{domain generalization} problem, which aims to build models that can perform well on a target domain which is sampled from a different distribution as compared to the source domain distribution(s). Successfully solving the domain generalization problem requires learning domain-invariant feature representations that can generalize to unseen domains. \blfootnote{$^*$ Work done while the author was an intern at Salesforce Research. Correspondence to \texttt{isabela.albuquerque@emt.inrs.ca}.} 

Current approaches to solving the domain generalization problem in computer vision typically perform ERM on the source domains by training a feature extractor on all available data sources~\cite{li2017deeper,li2018learning,carlucci2019domain} with or without additional strategies that enforce regularization on the feature extractor with an aim of improving generalization on the target domain. A  majority of these methods start with a pretrained feature extractor on the ImageNet~\cite{deng2009imagenet} dataset, finetune the feature extractor on all-but-one datasets from a dataset collection containing much fewer samples such as VLCS~\cite{torralba2011unbiased} and PACS~\cite{li2017deeper}, and evaluate the domain generalization performance on the held-out domain.  VLCS consists of PASCAL VOC 2007, LabelMe, Caltech101, SUN datasets with a total of 10729 samples and  PACS consists of Photos, Art Paintings, Cartoon, and Sketches datasets with a total of 9991 samples. These datasets, considered as unseen domains, present substantial similarity to ImageNet in that they contain images with very similar class labels. This makes the domain generalization problem easier. Moreover, supervised pretraining with ImageNet (or indeed any large scale supervised dataset) may lead to the network encoding strong class-discriminative biases for shapes~\cite{kriegeskorte2015deep} and textures~\cite{geirhos2018imagenet} on the pretraining dataset that may not be useful (or even hinder) domain generalization on other domains. 

An attractive alternative to using pretrained feature representations obtained from discriminative learning on datasets like ImageNet is to utilise unsupervised feature representation learning or Self-Supervised Learning (SSL). SSL aims to learn representations from unlabeled data by training feature encoders using pretext tasks---tasks that do not require per-sample human-annotated labels. 
For example, the Rotation task~\cite{gidaris2018unsupervised} trains a neural network to predict the degree of rotation of an image. Feature representations obtained from SSL can come close to or even match~\cite{He_2019_moco} the performance of  supervised learning methods on tasks such as image classification, object detection, and semantic segmentation. These feature representations have also been shown to improve adversarial robustness and out-of-distribution detection for difficult, near-distribution examples~\cite{hendrycks2019using}. 

In this paper, we show that a feature extractor trained with SSL can match or exceed the performance of a fully-supervised feature extractor on the domain generalization task. Specifically, multi-task SSL---combined training of multiple self-supervision pretext tasks---is able to learn feature representations that are robust to out-of-domain samples. Experiments on PACS and VLCS dataset show that SSL perform substantially better than supervised learning on datasets such as LabelMe and Sketch that represent a significant domain shift from ImageNet. On these datasets, models finetuned from multi-task self-supervised feature representations are better at localizing objects from the class of interest, as compared to supervised learning. Moreover, our method can be combined with other domain generalization algorithms, like invariant risk minimization, to obtain further performance improvement. 
In summary, self-supervised learning has the potential to outperform fully supervised learning for training deep learning algorithms that adapt to out-of-distribution data.  

\section{Related work}
\vspace{-0.3cm}
Self-supervised learning, as a form of unsupervised learning, aims to train a feature encoder from unlabeled data such that the learnt encoder is transferable to other downstream tasks. The training process usually involves solving a ``pretext" task with the purpose of learning good feature representations.
Example pretext tasks include image inpainting~\cite{Pathak_2016_inpaiting}, colorization~\cite{Zhang_2016_color,Zhang_2017_splitbrain}, prediction of patch orderings~\cite{Doersch_2015_context,Noroozi_2016_jigsaw} or rotation degree~\cite{gidaris2018unsupervised}. Some pretext tasks assign pseudo-labels to images by clustering~\cite{caron2018deep,Caron_2019_cluster}. Other pretext tasks train the encoder to discriminate instances by forming contrastive loss functions~\cite{Wu_2018_Instance,Ye_2019_end2end,Oord_2018_CPC,He_2019_moco}. Doersch and Zisserman~\cite{Doersch_2017_multitask} show that combining multiple pretext tasks with an architecture that uses a lasso technique for factoring representations leads to performance improvement over single tasks on image classification, object detection, and depth prediction tasks. Moreover, deep encoders trained with SSL can improve robustness to  adversarial or corrupted samples~\cite{hendrycks2019using} and improve few-shot learning~\cite{gidaris2019boosting,su2019does}.

Out-of-distribution generalization has been addressed by previous work under different settings. The domain adaptation literature focuses on strategies aimed at learning features capable of performing well under domain shift. Examples include Unsupervised Domain Adaptation \cite{ben2010theory}, which assumes that unlabeled samples from target domain are available during training. The target data can be used, for example, to adapt the learnt features on the source domain to reduce the mismatch between source and target domains~\cite{ganin2016domain}. A more general setting for out-of-distribution generalization consists of learning representations which are not adapted to a specific target domain. This is commonly referred as domain generalization and, in this case, no unlabeled target samples are assumed to be available at training time. Several recent efforts have addressed this problem by learning representations invariant to data distributions \cite{li2018domain}, incorporating domain shifts at training time \cite{li2018learning,dou2019domain}, or using data augmentation \cite{volpi2018generalizing}.

Recent work has adopted SSL to enforce the representation spaces learnt by neural networks to generalize to out-of-distribution data. The most pertinent related work for our paper is Carlucci et al. \cite{carlucci2019domain} who combine a discriminatory loss for supervised learning with an auxiliary loss for solving jigsaw puzzles, an SSL task. Zhai et al. \cite{zhai2019visual} also study the impact of self-supervision on learning transferable features, focusing on the performance of individual SSL tasks on classification tasks that may not have the same label space.  In this work, we show that carefully selected combination of self-supervised learning tasks trained with standard optimization techniques obtain comparable or better performance to supervised learning in the domain generalization setting.

\section{Methods}
\subsection{Problem Setting}
Let $\mathcal{X}$ and $\mathcal{Y}$, represent the data and label space, respectively. A domain $\mathcal{D}$ is defined as a joint probability distribution over $\mathcal{X} \times \mathcal{Y}$. We consider  a training set constructed by sampling pairs $(x^m, y^m) \sim \mathcal{D}_{S_i}$ from $N$ different source domains $\mathcal{D}_{S_i}$, and a test set $(x^{m'}, y^{m'})$ sampled from a target domain $\mathcal{D}_T$ distinct from all $\mathcal{D}_{S_i}$, $i=1:N$. We are interested in learning representations that generalize to unseen target domains, while employing examples only from the source domains at training time. Specifically, we tackle the \textit{homogeneous} domain generalization setting \cite{li2019episodic}, where all the domains share the label space $\mathcal{Y}$, i.e., the same classes are found across the source and target domains.  We note that this problem is fundamentally different from the popular unsupervised domain adaptation setting \cite{ben2010theory}, where the representation space is adapted to yield good performance for a specific target domain with unlabeled data sampled from this distribution.

\begin{figure}[t]
    \centering
    \includegraphics[width=\columnwidth]{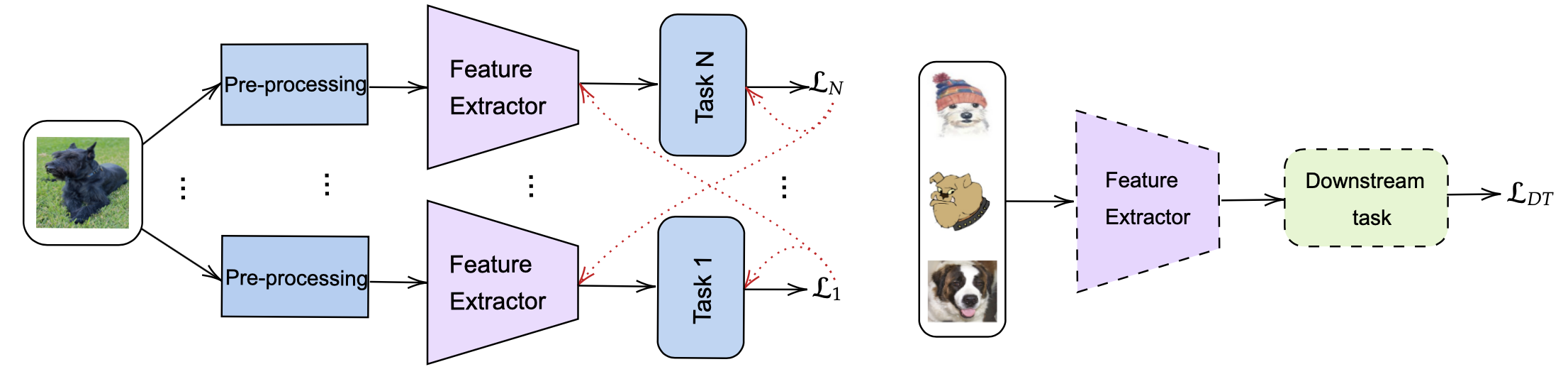}
    \caption{Illustration of the training scheme. \textbf{Left}: Self-supervised pretraining with multiple tasks.  The feature extractor is shared and is updated through the loss of all tasks. \textbf{Right}: Supervised finetuning for the domain generalization.}
    \label{fig:approach}
\end{figure}

\subsection{Self-Supervised Learning for Domain Generalization}
Our SSL approach for out-of-distribution generalization consists of two main steps: i) self-supervised pretraining, and ii) supervised fine-tuning. This setup differs from Carlucci et al.~\cite{carlucci2019domain} who finetune a representation learnt in a supervised manner using both supervised and self-supervised tasks simultaneously.

In our method, we use a feature extractor $F$ with parameters $\phi$ is responsible for encoding the input image. We feed the encoded feature representation to a model $T$ with parameters $\omega$ responsible for performing a specific self-supervised task. If $K$ tasks are considered at training time, we use $K$ task-specific modules denoted by $T_j$, $j=1:K$, with parameters $\omega_i$.  We perform preprocessing steps necessary for each task, encode the corresponding inputs, and feed the inputs to the corresponding task-specific module. We consider the $K$ losses provided by each $T_j$ to update the feature extractor by using the average across losses provided by each task-specific module as loss function. When there is no trade-off between optimizing the feature extractor for different tasks and the sample complexity for each task is reasonably similar, this approach is intuitively able to encode the input to an useful representation space for all tasks. Each $T_j$ is updated taking into account solely the loss corresponding to the $j$-th task (Figure~\ref{fig:approach}-Left). In the case where different tasks are expected to converge at different rates, we sequentially train the feature extractor on different tasks, by fine-tuning the model obtained on one task using another task. 


After updating $\phi$ on the self-supervised tasks, we feed encoded input and outputs class probabilities for the downstream task to a model $D$ with parameters $\theta$. If $N$ source domains are available at training, we find the optimal values of $\phi$ and $\theta$, denoted by $\phi^*$ and $\theta^*$ respectively, by performing ERM over all source domains:\vspace{-5pt}
\begin{equation}\label{ERM}
    \phi^*, \theta^* = \arg \min \frac{1}{N} \sum_{j=1}^{N} \ell(D(F(x_i)),y_i).
\end{equation}
Note that $\phi$ is updated in both self-supervised and supervised fine-tuning.

\begin{figure}[t]
    \centering
    \includegraphics[width=0.9\columnwidth]{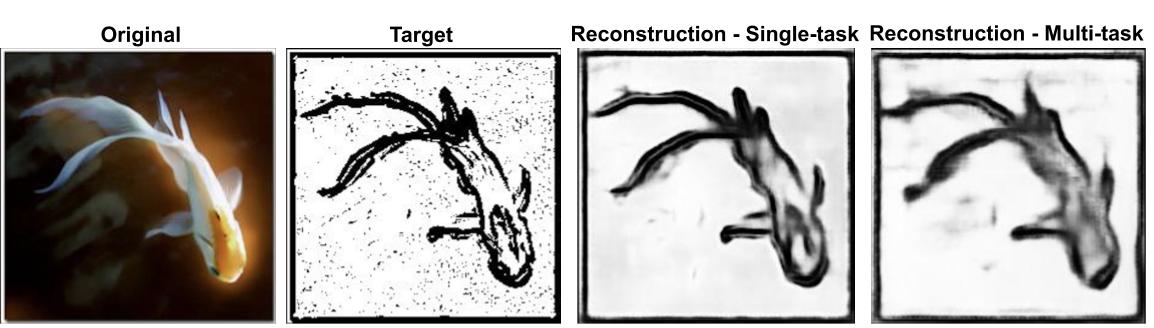}
    \caption{Gabor filter response reconstruction task. \textbf{Left}: Prediction by a model trained with the Gabor filter response reconstruction task alone.  \textbf{Right}: Prediction by a model simultaneously trained with DeepCluster, Rotation, and the Gabor filter response reconstruction task.} 
    \label{fig:pretext_G}
\end{figure}

\subsection{Pretext Tasks}
We now describe the SSL pretext tasks employed in this paper, including a novel Gabor filter response reconstruction task. 


\vspace{-5pt}\subsubsection{Gabor Filter Response Reconstruction:}
A Gabor filter is a two-dimensional spatial linear filter which highlights lower-level features in an image such as edges in a specific direction and texture \cite{fogel1989gabor}. 
Gabor filters are known to have similar properties as visual cortical cells of mammalian brain~\cite{daugman1980two,daugman1985uncertainty}. We are specifically interested in designing an SSL task based on Gabor filters to leverage their ability to capture low-level visual information. We can combine this task with SSL approaches that try to capture low/mid-level visual information, such as rotation prediction~\cite{gidaris2018unsupervised}, and high-level visual information, such as DeepCluster~\cite{caron2018deep}. 

Our proposed task is to train an encoder-decoder model given an input image, reconstruct the response of a Gabor filter bank considering seven distinct directions. We expect that, by learning to reconstruct the filter bank, the model will learn to capture the low-level features captured by the series of Gabor filters. In order to highlight the detected edges and to discourage the model from focusing on fine-grained information contained in the image, we subtract the original input from the filter bank response, convert it gray-scale, and apply a binary threshold on the intensity values of each pixel. The average pixel-wise binary cross-entropy between predicted and ground-truth filter responses is used as loss to update the parameters of the encoder-decoder model. 

\vspace{-5pt}\subsubsection{Rotation:} Gidaris et al.~\cite{gidaris2018unsupervised} proposed the rotation task which learns representations by training a model to predict the angle by which the input image is rotated. The authors argue that for a model to successfully learn to predict the angle of rotation, it needs to be able to capture information regarding pose, location, orientation, and the type of object present in the input image, as well as recognizing and localizing salient object parts in the image. Therefore, this task can be understood as taking into account both low-level features, such as orientation, as well as higher-level information, as the object type. 
The rotation task also forces the model to reduce the photographer bias~\cite{feng2019self}, making it easier to transfer these features to real-world tasks. 
As in \cite{gidaris2018unsupervised}, we consider four rotation angles to be predicted: $0^o, 90^o, 180^o$, and $270^o$.

\vspace{-5pt}\subsubsection{DeepCluster:}
The DeepCluster task~\cite{caron2018deep} learns a feature representation by training a model to predict clustering assignments to each data point. At the beginning of each epoch, the training data is clustered in the current representation space using $k$-means and the labels are then re-assigned according to which cluster each data point belongs to. Convolutional layers implementing Sobel filters are employed in the model input in order to remove color information and encourage the model to capture features such as edges and shape.  

\subsection{Architecture Details}\label{sec:arch}
Following prior work \cite{li2017deeper,li2018domain,carlucci2019domain,caron2018deep,gidaris2018unsupervised}, we utilize architectures based on AlexNet~\cite{krizhevsky2012imagenet} as the main backbone for all experiments. We replace instance normalization layers by batch normalization layers. When training a model with only Rotation and/or Gabor Reconstruction tasks, we decrease the number of filters on the first two convolutional blocks of AlexNet from 96 to 64 to match the architecture proposed by Gidaris et al.~\cite{gidaris2018unsupervised}. For all tasks, we consider as representation the output of the last convolutional block, which outputs a tensor of shape $(256, 6, 6)$, yielding a representation of size $9216$ after flattening. Next, we describe the architecture details for the task-specific heads and for the downstream domain generalization task.

\vspace{-5pt}\subsubsection{Rotation:} For the Rotation task, the angle of rotation for the input image is predicted by a fully-connected (FC) architecture that follows the design of the classifier head from AlexNet: $Dropout(0.5)\rightarrow FC(9216, 4096)\rightarrow ReLU\rightarrow Dropout(0.5)\rightarrow FC(4096,4096)\rightarrow ReLU\rightarrow FC(4096, 4)$. \vspace{-10pt}
\vspace{-5pt}\subsubsection{Gabor Reconstruction:} For reconstructing the Gabor filter bank response using a $9216$-size representation, we utilize as decoder an architecture designed to be a mirrored version of the AlexNet encoder. We replace the convolutional layers by transposed convolutions with the same parameters, except for the last convolutional layer, which maps the 64 channels of the input to a single-channel output, since we consider gray-scale filter responses.\vspace{-10pt}
\vspace{-5pt}\subsubsection{DeepCluster:} The task-specific head for DeepCluster was implemented following the design by Caron et al.~\cite{caron2018deep}, which is identical to the architecture for the rotation task head, except for the last FC layer that of size $(4096, n_c)$, where $n_c$ corresponds to the number of clusters. We set $n_c$ to 10000, following~\cite{caron2018deep}. This layer is re-initialized at the beginning of each epoch, when the clusters assignments are recomputed.\vspace{-10pt}
\vspace{-5pt}\subsubsection{Domain Generalization:} Following~\cite{kolesnikov2019revisiting}, we employ a model composed by a single FC layer mapping the representation from $9216$ to the number of classes specific to the domain generalization dataset. 

\section{Experiments}
To demonstrate that multi-task SSL is useful for achieving domain generalization, we perform four experiments which help answer the following questions: 1) How well does each model perform on each pretext task, and how is this performance affected by combining multiple tasks; 2) Are the representations learnt with SSL able to generalize to different domain shifts and which tasks are better suited for this goal; 3) Are the features learnt with SSL able to transfer across domains; and 4) What is the impact on out-of-distribution generalization when the sample diversity across the source domains is reduced. 

\subsection{Pretext Tasks}
In this set of experiments we evaluate the performance of individual tasks and of combinations of tasks. We combine tasks using two different approaches: 1) Average (AVG): The feature extractor parameters are updated with the aim of minimizing an average of the normalized losses provided by each task individually; 2) Fine-tuning (FT): The feature extractor is trained with one task until the task converges, then this task is dropped and a new task is introduced. \vspace{-5pt}

\vspace{-5pt}\subsubsection{Implementation Details:} We train each self-supervised model using the training partition of the ILSVRC 2012 datset and evaluate its performance on the validation partition to select hyperparameters. For all the tasks we use the Stochastic Gradient Descent (SGD) optimizer with Polyak's acceleration coefficient equal to 0.9. When training models with Rotation and Gabor Reconstruction tasks, we set the learning rate to 0.01, employ weight decay regularization with value 0.00005, and set the training budget to 20 epochs. The learning rate is decreased by a factor of 0.1 each 10 epochs. For DeepCluster, we perform experiments with the pretrained AlexNet released by the authors\footnote{\url{https://github.com/facebookresearch/deepcluster}} and use the same hyperparameters---a learning rate of 0.05 and weight decay of 0.00001. 

\begin{table}[t]
\caption{Performance of self-supervised models on pretext tasks. We measure accuracy for rotation task and reconstruction loss for the Gabor filter task. R: Rotation, G: Gabor, DC: DeepCluster. AVG: Models trained with average loss across tasks, FT: Models trained by sequentially finetuning tasks.}
\centering
\begin{tabular}{l|c|c|c|c|c}
\hline
& R & G & R+G (AVG) & R+DC (FT)   & R+G+DC (FT) \\ \hline
Rotation Accuracy & 90.19        & ---   & 87.45    & 72.68   & 79.92   \\
Gabor Loss  &   ---          & 0.42   & 0.46   & ---   & 0.48        \\ \hline
\end{tabular}
\label{tab:pretext}
\end{table}
\vspace{-5pt}\subsubsection{Pretext Task Performance:} We report the performance of the models trained on individual tasks as well as results obtained by combining different tasks. For the Rotation task, we report the average accuracy on the validation set as metric after 20 epochs. 
Table~\ref{tab:pretext} shows the performance of different models on the respective pretext tasks used at training time. The accuracy for Rotation decreases only slightly when  Gabor filter response reconstruction task is added, indicating that there is no strong conflict between those two tasks. When fine-tuning the DeepCluster model on Rotation, we observe a large drop on accuracy, indicating that the features obtained with the DeepCluster task do not present a good initialization for Rotation. When the Gabor reconstruction task is included in training, the accuracy obtained on Rotation increases more than $7\%$, showing a synergy between the two tasks. Finally, the Gabor filter task is helped by adding the higher level tasks: Rotation and DeepCluster.


\subsection{Domain Generalization}
We now describe the performance of feature representations obtained from the SSL tasks on VLCS and PACS, the two main domain generalization benchmarks. Each of these datasets are composed of four different datasets with the same classes. In all experiments, we fine-tune the self-supervised learnt representations using a leave-one-domain-out scheme, i.e.  the model is fine-tuned on the training examples from three domains and the best performance in terms of accuracy obtained on the unseen remaining domain is reported.

To isolate the effect of the type of pretraining, we use the same architecture (AlexNet) for the feature extractor  and the same architecture (a 1-layer neural network described in Section \ref{sec:arch}) for task head modules in all the evaluated strategies. 
We train all models for 100 epochs using SGD with learning rate equal to 0.001, Polyak's acceleration coefficient of 0.9, and weight decay regularization of 0.00005. We evaluate performance by computing the best accuracy achieved on the unseen target domain.\vspace{-5pt}

\begin{figure}[t]
\begin{minipage}[t]{0.49\linewidth}
\centering
\includegraphics[width=0.9\columnwidth]{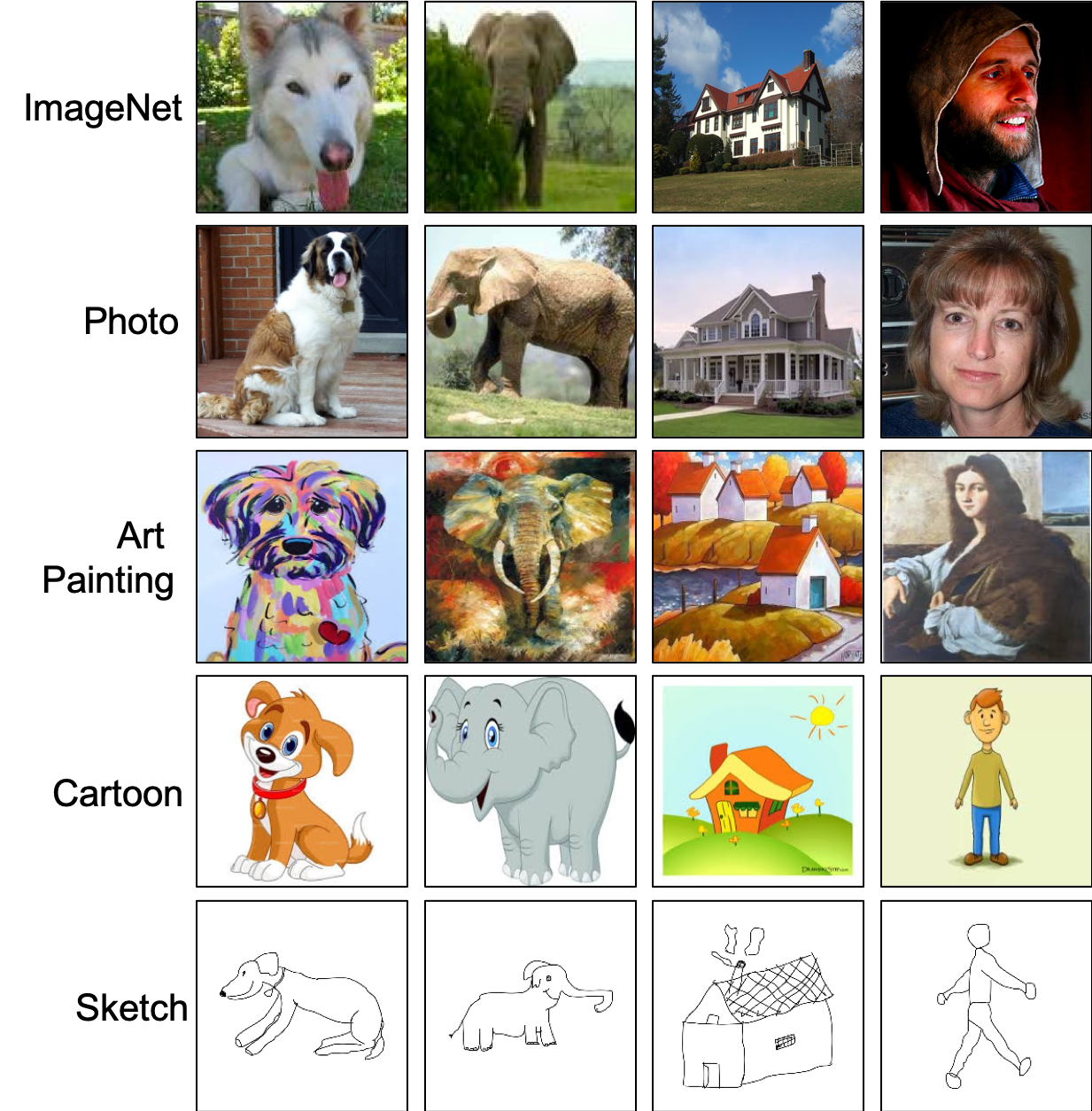}
\caption{Examples: ImageNet and PACS.}
\label{fig:example_pacs}
\end{minipage}
\hspace{0.5cm}
\begin{minipage}[t]{0.49\linewidth}
\centering
\includegraphics[width=0.9\columnwidth]{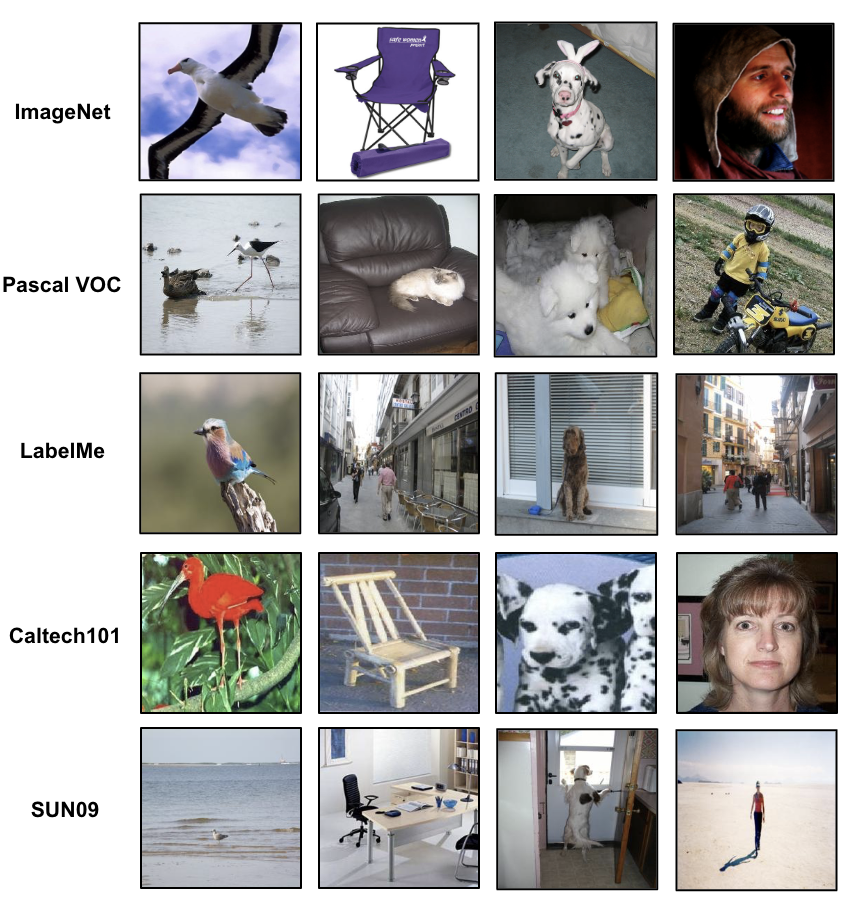}
\caption{Examples: ImageNet and VLCS.}
\label{fig:example_vlcs}
\end{minipage}
\end{figure}

\vspace{-5pt}\subsubsection{Baselines:} The performance of the representations learnt by self-supervision is compared with two baselines: a randomly initialized model and a model pretrained using the full training partition of the ILSVRC 2012 (ImageNet) dataset. Previous work on domain generalization has shown that fine-tuning a pretrained model on ImageNet on all source domains is a strong baseline for comparing the capability of generalizing to unseen domains. This is primarily because most of the datasets considered as unseen domains actually have considerable similarity with ImageNet, in that they contain natural images with classes overlapping with ImageNet. We do not include comparisons with methods such as \cite{li2018domain,li2018learning,carlucci2019domain} which use different stopping criteria, architectures, or combinations of loss functions for evaluating domain generalization performance and are hence not directly comparable. A comparison with these methods is included in the supplement.


\begin{figure}[b]
\centering
\captionof{table}{Domain generalization performance on the PACS benchmark. Multi-task self-supervised learning outperforms supervised learning on PACS. Accuracy reported in percent. Bolded value indicates best model for the target domain.}
\begin{tabular}{|c|cccccc|cc|}
\hline
\backslashbox{Domain}{Method} & R & G & DC & R+G & R+DC  & R+G+DC & Rand. Init. & Supervised \\ \hline
P & 80.96 & 77.66 & 79.88 & 82.28 &  85.99 & 84.31 & 70.12 & \textbf{87.19}  \\
A & 54.20 & 47.71 & 54.74 & 56.01 &  \textbf{62.65} & 61.67 & 45.21 & 61.67  \\
C & 65.10 & 58.62 & 62.29 & 65.61 &  62.97 & \textbf{67.41} & 53.58 & 64.85  \\
S & 63.76 & 55.61 & 44.18 & 60.45 &  60.73 & \textbf{63.91} & 53.50 & 55.61  \\ \hline
Average & 66.00 & 59.90 & 60.27 & 66.08 & 68.08 & \textbf{69.32} & 55.60 & 67.33 \\ \hline
\end{tabular}
\label{tab:dg_pacs}        
\end{figure}

\begin{figure}[t]
\centering
\captionof{table}{Domain generalization performance on the VLCS benchmark. Multi-task self-supervised learning performs comparably to supervised learning on VLCS.  Accuracy reported in percent. Bolded value indicates best model for the target domain.}
\begin{tabular}{|c|cccccc|cc|}
\hline
\backslashbox{Domain}{Method} & R & G & DC & R+G & R+DC  & R+G+DC & Rand. Init. & Supervised \\ \hline
V & 60.41 & 53.31 & 61.20 & 57.95 & {62.59} & 57.65 & 51.14 & \textbf{64.07}  \\
L & \textbf{66.12}  & 61.86 & 59.85 & 65.87 & 62.86 & 64.99 & 59.22 & 60.73  \\
C & 84.20 & 78.77 & {94.10} & 87.97 & 93.87 & 89.15 & 74.06 & \textbf{95.52}  \\
S & 59.70 & 56.95 & 57.66 & 59.09 &{59.80} & 58.88 & 55.03 & \textbf{62.44}  \\ \hline
Average & 67.60 & 62.73 & 68.20 & 67.72 & {69.78} & 67.67 & 59.86 & \textbf{70.69} \\ \hline
\end{tabular}
\label{tab:dg_vlcs}        
\end{figure}

\vspace{-5pt}\subsubsection{PACS:} The PACS benchmark  was proposed as a test bed for out-of-distribution generalization strategies that presents a high overall domain shift from ImageNet~\cite{li2017deeper}. 
PACS contains four domains: Photo, Art painting, Cartoon, and Sketch (Figure \ref{fig:example_pacs}). Each dataset is divided into seven classes: dog, elephant, giraffe, guitar, horse, house, and person. 
In Table \ref{tab:dg_pacs}, we show the performance of self-supervised learning methods, along with the baseline models obtained by supervised pretraining and random initialization. 

All single- and multi-task SSL approaches significantly outperform the randomly initialized baseline. 
As we combine multiple SSL tasks, the average performance for domain generalization improves. The SSL tasks complement each other, boosting the accuracy on the unseen domain by, for instance, $3.32\%$ when comparing Rotation with Rotation + Gabor + DeepCluster. Saliently, the combination of all three tasks surpass the performance of the supervised pretrained baseline by 2\% on average and are better than the supervised pretrained baseline on Art painting, Cartoon, and Sketch domains. The combination of Rotation and DeepCluster also outperforms the supervised pretrained baseline on average. 

SSL obtains significant improvement over supervised pretraining on the Art painting, Cartoon, and Sketch domains, which represent a  significant domain shift from natural images present in ImageNet. These results indicate that self-supervised tasks are able to learn a feature representation that is more readily transferable across domains as compared strongly discriminative supervised learning on the same set of images. 
Interestingly, self-supervised pretraining shows the highest performance improvement (8.3\% for R + G + DC) on the Sketch dataset over the supervised model. Images in the Sketch dataset contain, not surprisingly, simple sketches that lack texture and color (Figure \ref{fig:example_pacs}). The self-supervised learning approach, containing tasks such as Rotation and Gabor filter reconstruction that focus on low- and mid-level features, may allow the model to capture information related to edges and shapes without capturing texture information. 
Note that the model pretrained with DeepCluster alone performs the worst on Sketch dataset, but the performance is recovered once Rotation and Gabor filter reconstruction tasks are included, confirming the importance of adding low-level tasks to pretraining. 


\vspace{-5pt}\subsubsection{VLCS:} The VLCS benchmark \cite{fang2013unbiased} contains natural images obtained from the PASCAL VOC \cite{everingham2010pascal}, LabelMe \cite{russell2008labelme}, Caltech101 \cite{fei2004learning}, and SUN09 \cite{choi2010exploiting} datasets divided in five classes: bird, car, chair, dog, and person. Following convention, we split each dataset into training and validation sets that contain $80\%$ and $20\%$ of the data points, respectively. 

In Table \ref{tab:dg_vlcs}, we summarize the results of  single-task and multi-task self-supervised pretraining strategies with the randomly initialized and ImageNet-initialized models. The average performance of best multi-task self-supervised model (R + DC) across datasets (69.78\%) is significantly better than random initialization (59.86\%) and almost matches the fully-supervised model (70.69\%). Saliently, 5 out of 6 SSL strategies beat the performance of the fully-supervised model on the LabelMe datset. As Figure~\ref{fig:example_vlcs} shows, LabelMe represents a significant domain shift when compared with ImageNet; the objects are usually smaller in comparison to ImageNet and larger, distractor objects which do not belong to the class label are often  present in the image. The supervised model slightly outperforms SSL on PASCAL VOC,  Caltech101, and SUN09---datasets that are relatively more similar to ImageNet.
Among the self-supervised tasks, R + DC obtain the best overall performance, followed by R + G + DC. Some individual tasks such as DC obtain better performance than multi-task models like R + G and R + G + DC. However, unlike the PACS dataset, multi-task SSL does not uniformly improve the performance over individual tasks. 

\vspace{-5pt}\subsubsection{Qualitative Differences:} 
We perform a qualitative evaluation of the feature representations learnt by SSL and fully-supervised learning methods by visualizing the input regions that obtain the highest model activations for the predicted class, using the GradCAM heatmap method~\cite{selvaraju2017grad}. Specifically, we consider the R + G + DC pretraining method which outperformed the supervised pretrained model on PACS and closely matched the performance on VLCS. 

\begin{figure}[h!]
\centering
\begin{tabular}{ccccccccc}
\tiny{Original} & \tiny{R+G+DC} & \tiny{Supervised} & \tiny{Original} & \tiny{R+G+DC} & \tiny{Supervised} & \tiny{Original} & \tiny{R+G+DC} & \tiny{Supervised} \\
\multicolumn{1}{c}{\tiny{Dog}} \\
\includegraphics[width = 0.42in]{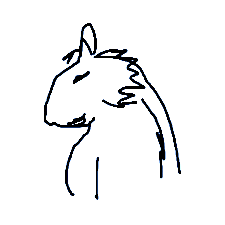} &
\includegraphics[width = 0.42in]{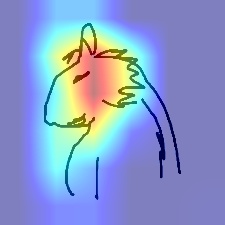} &
\includegraphics[width = 0.42in]{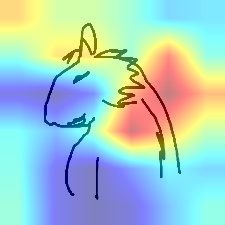} &
\includegraphics[width = 0.42in]{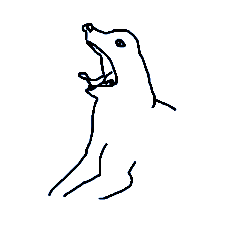} &
\includegraphics[width = 0.42in]{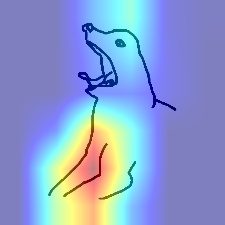} &
\includegraphics[width = 0.42in]{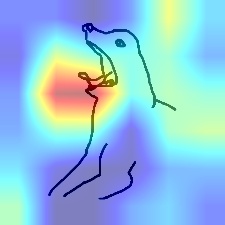} &
\includegraphics[width = 0.42in]{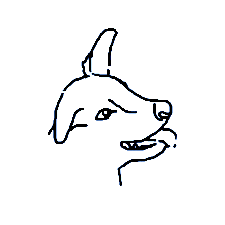} & 
\includegraphics[width = 0.42in]{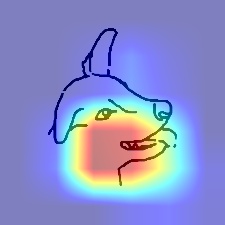} & 
\includegraphics[width = 0.42in]{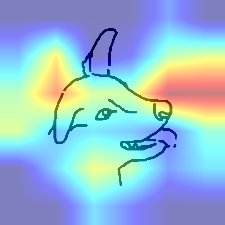} \\
\multicolumn{1}{c}{\tiny{Elephant}} \\
\includegraphics[width = 0.42in]{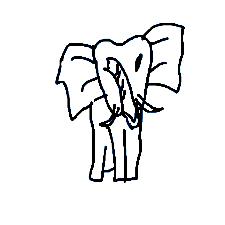} &
\includegraphics[width = 0.42in]{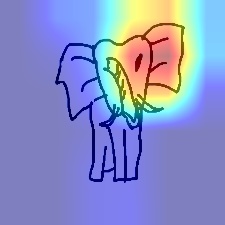} &
\includegraphics[width = 0.42in]{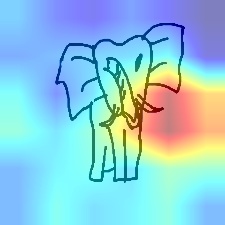} &
\includegraphics[width = 0.42in]{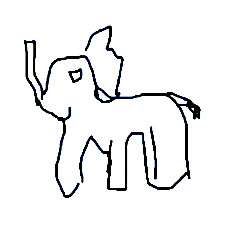} &
\includegraphics[width = 0.42in]{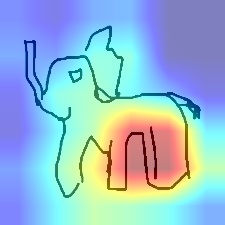} &
\includegraphics[width = 0.42in]{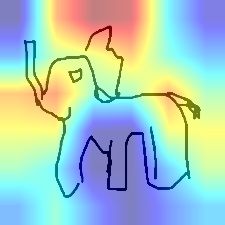} &
\includegraphics[width = 0.42in]{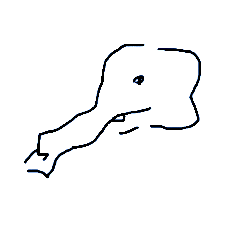} & 
\includegraphics[width = 0.42in]{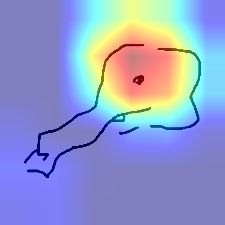} & 
\includegraphics[width = 0.42in]{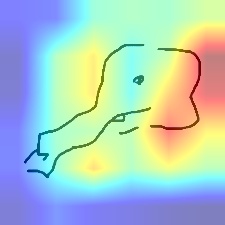} \\
\multicolumn{1}{c}{\tiny{Giraffe}} \\
\includegraphics[width = 0.42in]{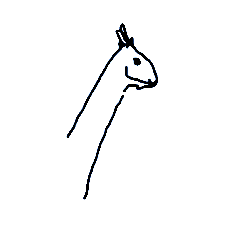} &
\includegraphics[width = 0.42in]{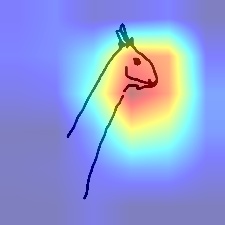} &
\includegraphics[width = 0.42in]{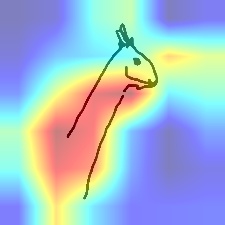} &
\includegraphics[width = 0.42in]{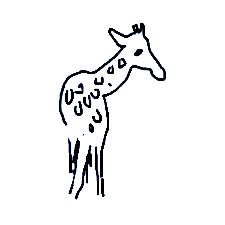} & 
\includegraphics[width = 0.42in]{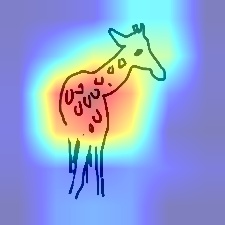} & 
\includegraphics[width = 0.42in]{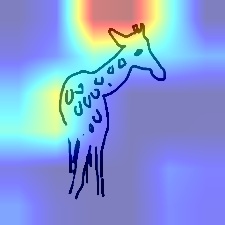} &
\includegraphics[width = 0.42in]{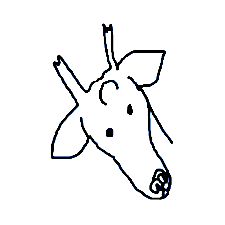} &
\includegraphics[width = 0.42in]{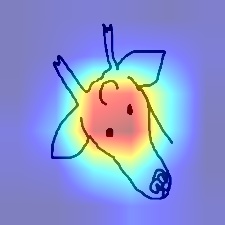} &
\includegraphics[width = 0.42in]{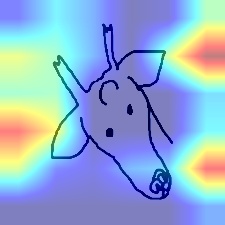} \\
\multicolumn{1}{c}{\tiny{Guitar}} \\
\includegraphics[width = 0.45in]{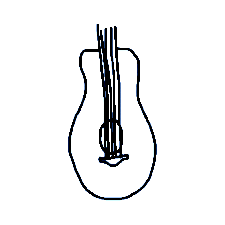} &
\includegraphics[width = 0.42in]{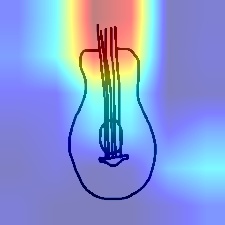} &
\includegraphics[width = 0.42in]{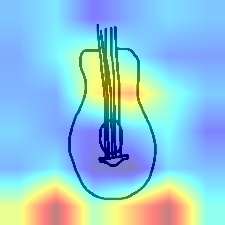} &
\includegraphics[width = 0.45in]{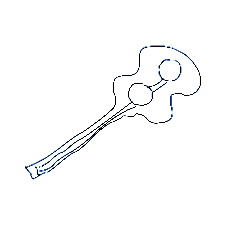} & 
\includegraphics[width = 0.42in]{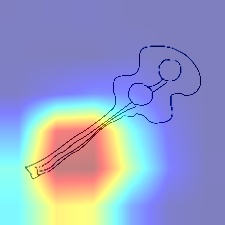} & 
\includegraphics[width = 0.42in]{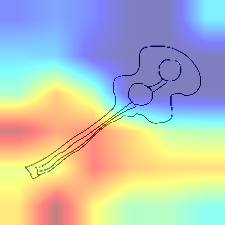} &
\includegraphics[width = 0.42in]{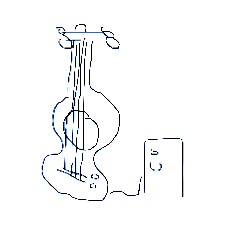} &
\includegraphics[width = 0.42in]{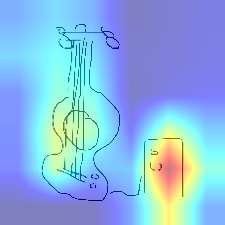} &
\includegraphics[width = 0.42in]{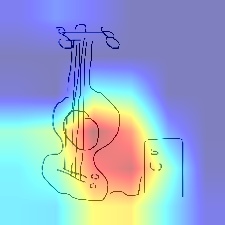} \\
\multicolumn{1}{c}{\tiny{Horse}} \\
\includegraphics[width = 0.42in]{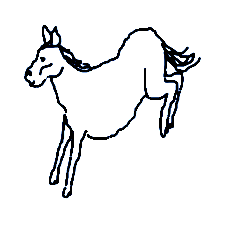}&
\includegraphics[width = 0.42in]{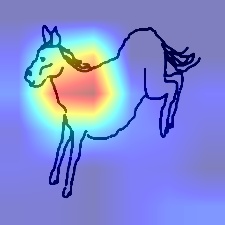} &
\includegraphics[width = 0.42in]{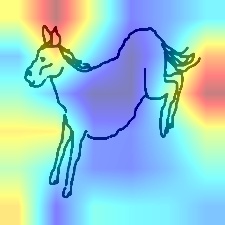} &
\includegraphics[width = 0.42in]{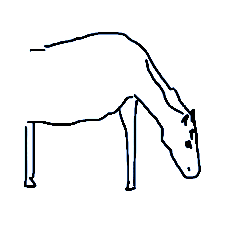} & 
\includegraphics[width = 0.42in]{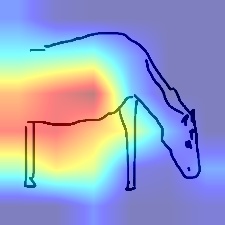} & 
\includegraphics[width = 0.42in]{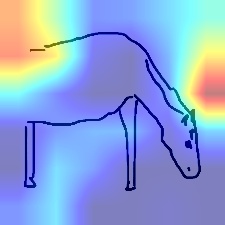} &
\includegraphics[width = 0.42in]{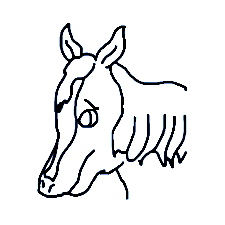} &
\includegraphics[width = 0.42in]{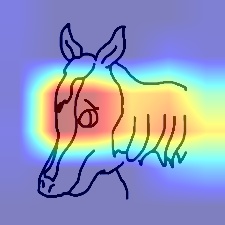} &
\includegraphics[width = 0.42in]{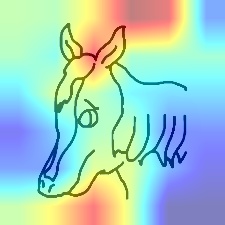} \\
\multicolumn{1}{c}{\tiny{House}} \\
\includegraphics[width = 0.45in]{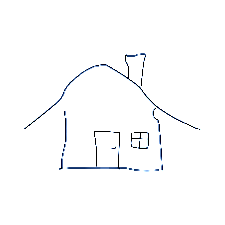}&
\includegraphics[width = 0.42in]{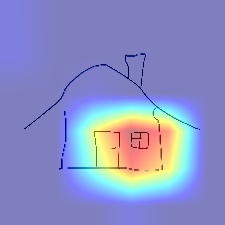} &
\includegraphics[width = 0.42in]{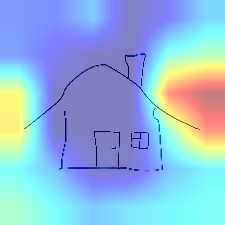} &
\includegraphics[width = 0.45in]{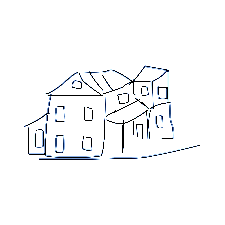} & 
\includegraphics[width = 0.42in]{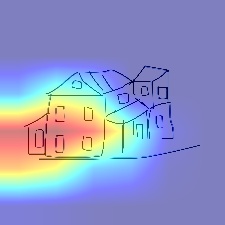} & 
\includegraphics[width = 0.42in]{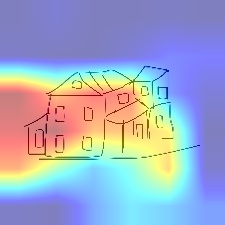} &
\includegraphics[width = 0.45in]{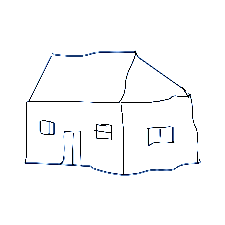} &
\includegraphics[width = 0.42in]{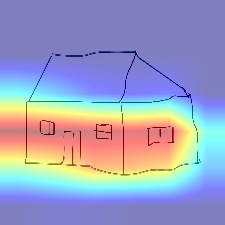} &
\includegraphics[width = 0.42in]{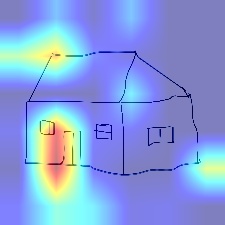} \\
\multicolumn{1}{c}{\tiny{Person}} \\
\includegraphics[width = 0.42in]{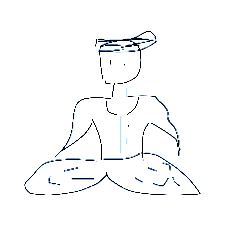}&
\includegraphics[width = 0.42in]{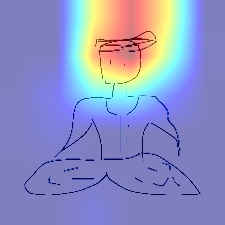} &
\includegraphics[width = 0.42in]{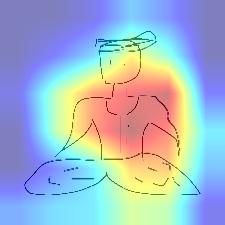} &
\includegraphics[width = 0.45in]{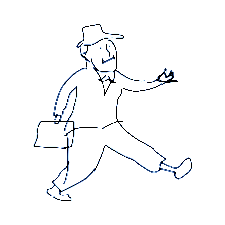} & 
\includegraphics[width = 0.42in]{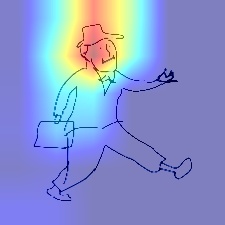} & 
\includegraphics[width = 0.42in]{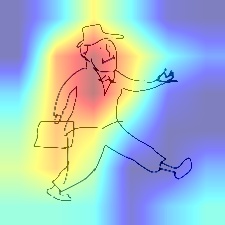} &
\includegraphics[width = 0.45in]{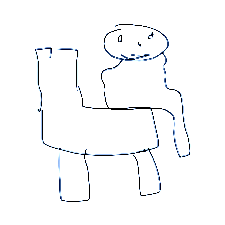} &
\includegraphics[width = 0.42in]{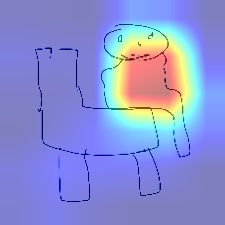} &
\includegraphics[width = 0.42in]{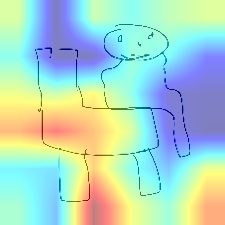} \\

\end{tabular}
\caption{GradCam visualizations for examples {correctly} classified by a model pretrained using Rotation, Gabor reconstruction, and DeepCluster (R+G+DC) and {misclassified} by the supervised baseline. Regions more relevant for prediction are shown in red. Models trained with self-supervision show better localization performance. Both models were fine-tuned on the PACS benchmark using Photo, Art painting, and Cartoon as source domains.}
\label{fig:gradcam}
\end{figure}

Figure \ref{fig:gradcam} shows the performance on the PACS benchmark with Sketch as target domain. Regions more relevant for prediction are shown in red.  Heatmaps corresponding to examples that were correctly classified by the self-supervised pretrained model and misclassified by the supervised baseline are shown along with the original input image. We observe that the multi-task self-supervised pretrained model is much better at focusing on parts of objects (such as heads and ears of animals, windows of houses), while ignoring the background. On the other hand, the supervised baseline considers larger portions of the input image for the prediction and frequently focuses on the background or distractor objects (e.g., the chair besides a person for the `person' class). 

\begin{figure}[h]
\centering
\begin{tabular}{ccccccccc}
\tiny{Original} & \tiny{R+G+DC} & \tiny{Supervised} & \tiny{Original} & \tiny{R+G+DC} & \tiny{Supervised} & \tiny{Original} & \tiny{R+G+DC} & \tiny{Supervised} \\
\multicolumn{1}{c}{\tiny{Bird}} \\ 
\includegraphics[width = 0.42in]{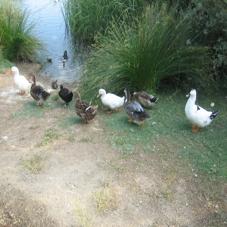} &
\includegraphics[width = 0.42in]{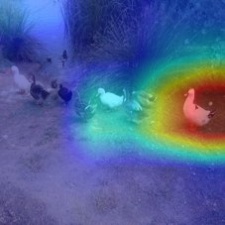} &
\includegraphics[width = 0.42in]{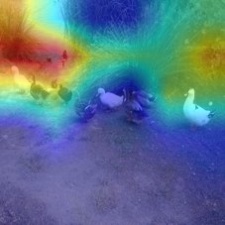} &
\includegraphics[width = 0.42in]{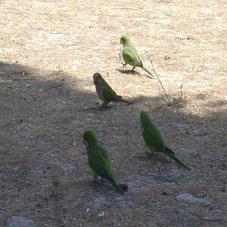} &
\includegraphics[width = 0.42in]{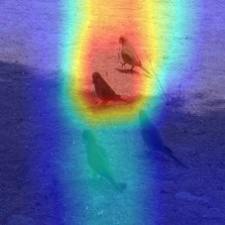} &
\includegraphics[width = 0.42in]{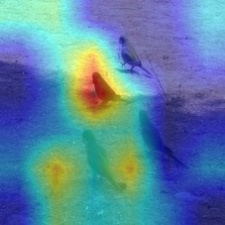} &
\includegraphics[width = 0.42in]{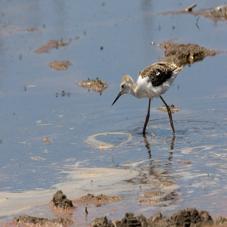} & 
\includegraphics[width = 0.42in]{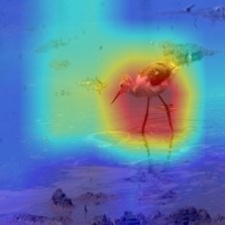} &
\includegraphics[width = 0.42in]{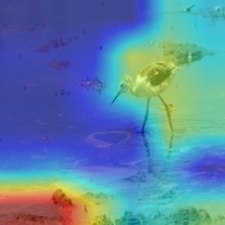} \\ 
\multicolumn{1}{c}{\tiny{Car}} \\ 
\includegraphics[width = 0.42in]{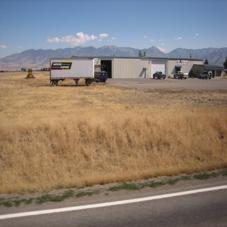} &
\includegraphics[width = 0.42in]{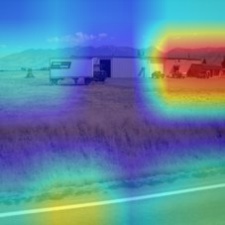} &
\includegraphics[width = 0.42in]{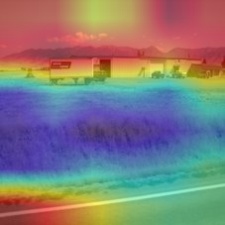} &
\includegraphics[width = 0.42in]{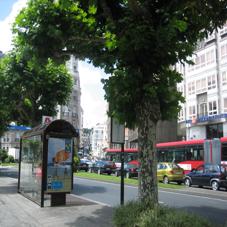} &
\includegraphics[width = 0.42in]{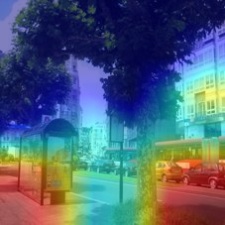} &
\includegraphics[width = 0.42in]{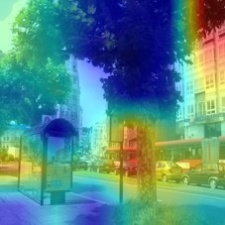} &
\includegraphics[width = 0.42in]{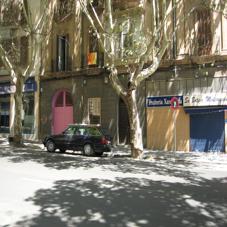} & 
\includegraphics[width = 0.42in]{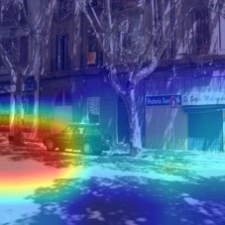} &
\includegraphics[width = 0.42in]{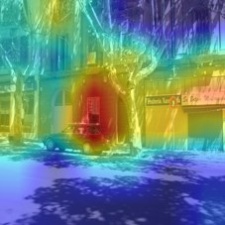} \\ 
\multicolumn{1}{c}{\tiny{Person}} \\ 
\includegraphics[width = 0.42in]{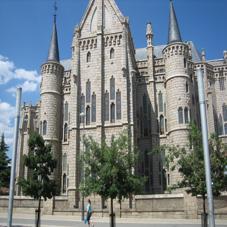} &
\includegraphics[width = 0.42in]{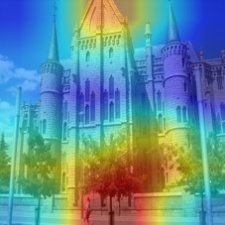} &
\includegraphics[width = 0.42in]{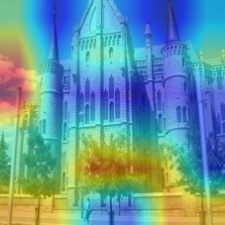} &
\includegraphics[width = 0.42in]{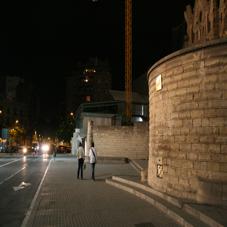} &
\includegraphics[width = 0.42in]{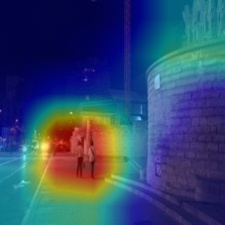} &
\includegraphics[width = 0.42in]{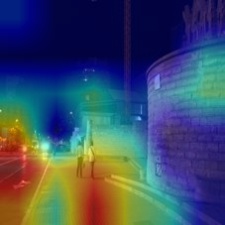} &
\includegraphics[width = 0.42in]{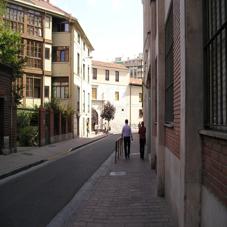} & 
\includegraphics[width = 0.42in]{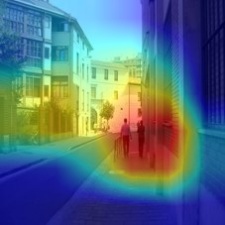} &
\includegraphics[width = 0.42in]{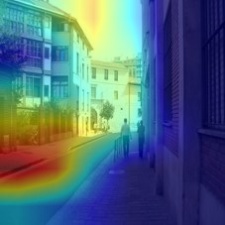} 

\end{tabular}
\caption{GradCam visualizations for examples {correctly} classified by a model pretrained using Rotation, Gabor reconstruction, and DeepCluster (R+G+DC) and {misclassified} by the supervised baseline. Regions more relevant for prediction are shown in red. Models trained with self-supervision show better localization performance. Both models were fine-tuned on the VLCS benchmark using Pascal VOC, Caltech101, and SUN09 as source domains. }
\label{fig:gradcam_vlcs}
\end{figure}

We observe similar trends on the VLCS benchmark with LabelMe as the target domain (Figure \ref{fig:gradcam_vlcs}), which contains natural images. The SSL model is much better at localizing small objects corresponding to the class of interest, while ignoring the background and distractor objects, for classes such as bird, car, and person. In contrast, the supervised baseline is more distracted by surrounding objects in the LabelMe dataset, which contains significantly more contextual information than ImageNet. 

\subsection{Cross-domain Transfer}
\begin{figure}[t]
    \centering
    \includegraphics[width=0.8\columnwidth]{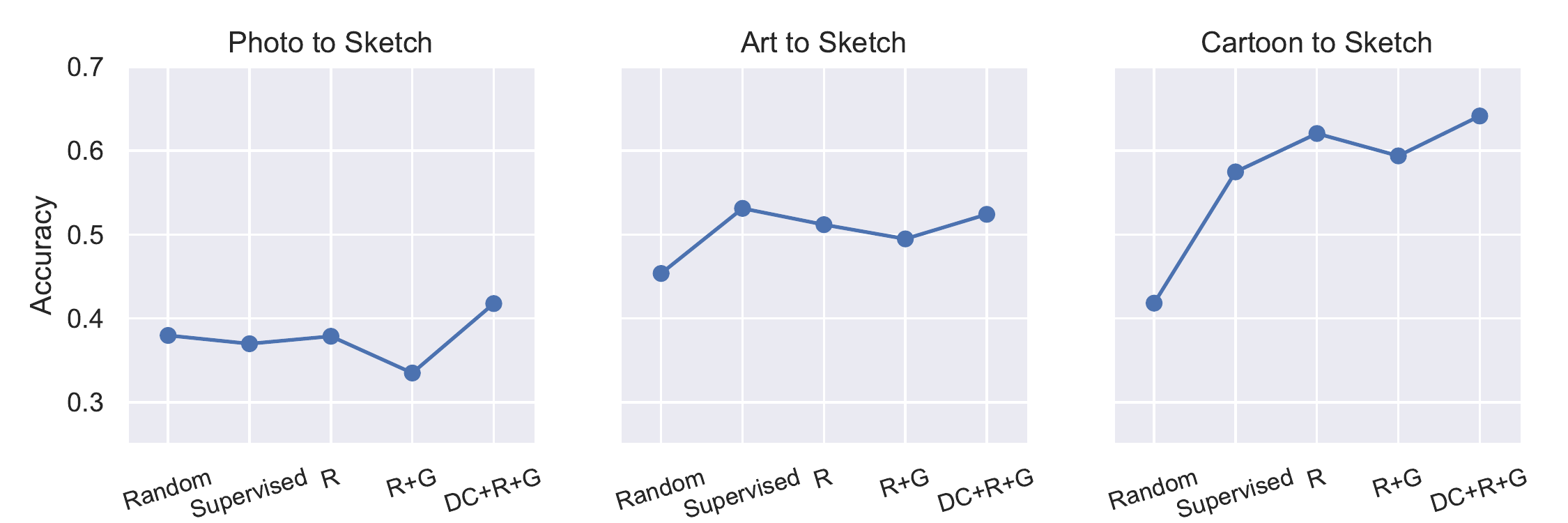}
    \caption{Performance for PACS cross-domain transfer using Sketch as target. Self-supervised learning with unlabeled images outperforms the supervised baseline.} 
    \label{fig:cross_sketch}
\end{figure}

\begin{figure}[t]
    \centering
    \includegraphics[width=0.8\columnwidth]{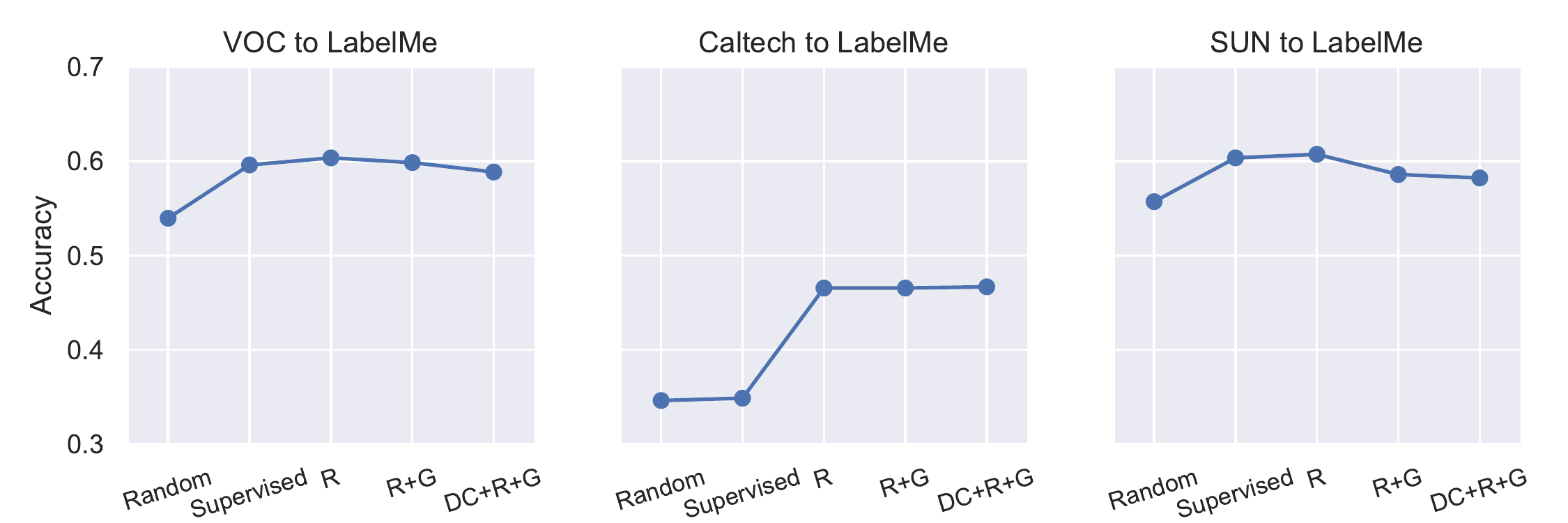}
    \caption{Performance for VLCS cross-domain transfer using LabelMe as target. Self-supervised learning with unlabeled images is comparable to, or outperforms, the supervised baseline.}
    \label{fig:cross_labelme}
\end{figure}

We also evaluate the performance of each representation on a cross-domain scenario where only one source domain is available at time, to investigate the case of low-data fine-tuning. For this purpose, we fine-tune each model with a training set composed of a single source domain and then evaluate the learnt representations on different domain. In the case of the VLCS benchmark, we perform experiments considering LabelMe as target and each remaining domain as source. Similarly, for the PACS benchmark, we use Sketch as target and each remaining domain as source. Results are shown in Figures \ref{fig:cross_labelme} and  \ref{fig:cross_sketch}. When the source domain datasets are similar to ImageNet (Caltech101 and Photo, for the VLCS and PACS benchmarks, respectively), the features learnt by models pretrained with self-supervised tasks yield better out-of-distribution generalization as compared to supervised learning. In other words, a neural network trained with roughly 1.2 million {unlabeled} images with self-supervised pretext tasks and finetuned with roughly 1500 labeled images obtains comparable or significantly better performance performance than a neural network with the same architecture trained on roughly 1.2 million {labeled} images and finetuned with approximately 1500 labeled images. These results also indicate that the self-supervision can be used to mitigate the effects caused by lack of visual diversity between the datasets employed in the pretraining and finetuning stages. 


\subsection{Combination with Other Domain Generalization Methods}
Finally, we study if a feature representation learnt with SSL can serve as good initialization for domain generalization methods that utilize different optimization techniques or loss functions to improve OOD performance. Specifically, we use Invariant Risk Minimization (IRM) \cite{arjovsky2019invariant}, a recently proposed method with strong performance. For these experiments, we finetune the pretrained model using IRM in order to enforce learning a representation for which the best predictor is the same across all the training domains. We use the same hyperparameters as in \cite{arjovsky2019invariant} for the colored MNIST experiments\footnote{\url{https://github.com/facebookresearch/InvariantRiskMinimization}}. A more extensive hyperparameter search is likely to improve performance across methods.

We find that combining IRM with SSL yields better out-of-distribution performance as compared to supervised learning for both PACS and VLCS benchmarks on average (Table \ref{tab:IRM}). For the PACS benchmark, SSL improves the best target accuracy on 3 out of  4  domains. For the VLCS benchmark,  SSL outperforms  the performance on 2 target domains, including Caltech101. Note that in  previous experiments (Table~\ref{tab:dg_vlcs}) with  ERM, supervised learning was superior to SSL on Caltech101. 
Overall, this experiment indicates that combining domain generalization strategies along with self-supervised pretraining can be an effective way to boost the out-of-distribution generalization capability of previously proposed methods.  

\begin{table}[t]
\centering
\caption{Performance of domain generalization with Invariant Risk Minimization (IRM). Self-supervised learning obtains better performance than supervised learning when using IRM for domain generalization. Bolded  value  indicates  best  model  for the target domain.}\vspace{3pt}
\resizebox{1\columnwidth}{!}{
\begin{tabular}{|c|cc|c|cc|}
\hline
\backslashbox{Domain}{Method}  & IRM-Supervised & IRM-R+G+DC & \backslashbox{Domain}{Method}  & IRM-Supervised & IRM-R+DC \\ \hline
P & \textbf{79.76} & 77.31  & V & \textbf{63.18} & 59.33 \\
A & 54.05 & \textbf{59.67}  & L & 59.10 & \textbf{62.11} \\
C & 61.43 & \textbf{63.78}  & C & 87.74 & \textbf{91.51} \\
S & 46.50 & \textbf{62.66}  & S & \textbf{61.01} & 60.91 \\ \hline
Avg. & 60.44 & \textbf{65.86}  & Avg. & 67.76 & \textbf{68.46} \\ \hline
\end{tabular}}
\label{tab:IRM}
\end{table}

\section{Conclusion}\vspace{-10pt}
Self-supervised learning has emerged as a powerful framework for learning feature representation that can match the performance of supervised learning on problems like image classification and few-shot learning. Here we show that feature representations obtained from self-supervised learning, especially those obtained by combining multiple pretext tasks, are able to match or exceed the performance of fully-supervised feature extractors on the domain generalization task and even improve localization. Moreover, self-supervision can be combined with other techniques that aim to learn feature representations which are amenable to domain generalization. Future work in this area can explore the performance of contrastive pretext tasks on domain generalization and alternate optimization strategies for training multi-task self-supervised learning models. 

%
%

\bibliographystyle{splncs04}
\bibliography{eccv2020submission}
\clearpage
\section*{Supplementary Material}
\subsubsection*{Comparing with domain generalization strategies:} For the sake of completeness, we compare the performance obtained by our best self-supervised pretrained models with previous work on domain generalization that proposes strategies for out-of-distribution generalization. These methods train models on top of model weigths pretrained with ImageNet. Even though the performance is not directly comparable across the different methods due to large differences on the architecture and training budget, we believe this comparison is valuable to show the gap between self-supervision for representation learning (which is not specifically designed for out-of-distribution generalization) and strategies that aim  to learn features robust to domain mismatches. We show in Tables S\ref{tab:vlcs_dg_prevwork} and  S\ref{tab:pacs_dg_prevwork} the results obtained by CIDDG \cite{li2018domain} and MLDG \cite{li2018learning} on the VLCS and PACS benchmarks, respectively. Since the  performance of both methods is reported in the literature by computing the performance on the target domain achieved by the model with best accuracy on the source domains, we also show in both tables the results achieved by the best self-supervised strategies under the same criterion. These results are indicated in the tables by the symbol $\dagger$. In addition, we  report the results obtained by the best self-supervised pretrained models on the target domain (indicated by $\ddagger$), as well as the performance achieved by JiGen \cite{carlucci2019domain}, which is also not directly comparable to the other results reported in the tables since the training stopping criterion was not specified. We also include in the tables the reported performance by the respective supervised baseline (denoted as DeepAll) for each method. Our best models are comparable to supervised methods trained with additional domain generalization techniques. 

\begin{figure}[h]
\begin{minipage}[t]{0.49\linewidth}
\centering
\captionof{table}{Comparison with previously report domain generalization performance on the VLCS benchmark.}
\centering
\resizebox{\columnwidth}{!}{
\begin{tabular}{cccccc}
\hline
              & V & L & C & S & Average \\ \hline
DeepAll-CIDDG$^\dagger$ & 62.71 & 61.28 & 85.73 & 59.33 & 67.26 \\
CIDDG$^\dagger$ & 64.38 & 63.06 & 88.83 & 62.10 & 69.72 \\
DeepAll-JiGen & 71.96 & 59.18 & 96.93 & 62.57 & 72.66 \\
JiGen         & 70.62  & 60.90 & 96.93 & 64.30 & 73.19 \\ \hline
R+DC$^\dagger$ & 62.19 & 59.10 & 87.74 & 58.58 & 66.90 \\
R+DC$^\ddagger$ & 62.59 & 62.86 & 93.87 & 59.80 & 69.78 \\ \hline
\end{tabular}
\label{tab:vlcs_dg_prevwork}
}
\end{minipage}
\hspace{0.1cm}
\begin{minipage}[t]{0.49\linewidth}
\centering
\captionof{table}{Comparison with previously reported domain generalization performance on the PACS benchmark.}
\centering
\resizebox{\columnwidth}{!}{
\begin{tabular}{cccccc}
\hline
              & P & A & C & S & Average \\ \hline
DeepAll-MLDG$^\dagger$ & 86.67 & 64.91 & 64.28 & 53.08 & 67.24 \\
MLDG$^\dagger$ & 88.00 & 66.23 & 66.88 & 58.96 & 70.01 \\
DeepAll-JiGen & 89.98 & 66.68 & 69.41 & 60.02 & 71.52 \\
JiGen         & 89.00 & 67.63 & 71.71 & 65.18 & 73.38 \\ \hline
R+G+DC$^\dagger$ & 84.31 & 61.67 & 67.41 & 57.47 & 65.18 \\
R+G+DC$^\ddagger$ & 84.31 & 61.67 & 67.41 & 63.91 & 69.32 \\ \hline

\end{tabular}
\label{tab:pacs_dg_prevwork}
}   
\end{minipage}
\end{figure}

\subsection*{Details of Gabor filter bank hyperparameters:} We implement the Gabor filter bank using the OpenCV-Python toolbox getGaborKernel function with the following hyperparameters: 
\begin{itemize}
    \item Kernel size: 10
    \item $\theta$: $0, \frac{\pi}{8}, \frac{\pi}{4}, \frac{\pi}{2}, -\frac{\pi}{8}, -\frac{\pi}{4}, -\frac{\pi}{2}$
    \item $\lambda$: 10
    \item $\sigma$: 4
    \item $\gamma$: 0.5
    \item $\Psi$: 0.0 
\end{itemize}

\end{document}